\pdfoutput=1

\documentclass[11pt]{article}
\usepackage[preprint]{acl} 

\usepackage{times}
\usepackage{latexsym}
\usepackage{subcaption}
\usepackage[many]{tcolorbox}
\usepackage{colortbl}
\usepackage{changepage}
\definecolor{blond}{rgb}{0.98, 0.94, 0.75}
\definecolor{lightgreen}{rgb}{0.5, 1.0, 0.83}
\definecolor{pastelyellow}{rgb}{0.99, 0.99, 0.59}
\definecolor{pastelorange}{rgb}{1.0, 0.7, 0.28}
\usepackage{caption}
\usepackage{float}
\usepackage{soul}
\usepackage{xcolor}
\usepackage[flushleft]{threeparttable}
\usepackage{enumitem}
\usepackage[T1]{fontenc}

\usepackage[utf8]{inputenc}
\usepackage{newunicodechar}
\newunicodechar{̇}{\.{}}
\usepackage{microtype}
\usepackage{inconsolata}
\usepackage{graphicx}

\title{A Comparative Study of Quality Evaluation Methods for Text Summarization}

 \author{Huyen Nguyen, Haihua Chen, Lavanya Pobbathi, Junhua Ding\textsuperscript{*}  \\
        University of North Texas, Denton, TX \\ 
         \texttt{huyennguyen5@my.unt.edu}, \texttt{haihua.chen@unt.edu},\\ \texttt{lavanyapobbathi@my.unt.edu},\texttt{junhua.ding@unt.edu}
         \\
  \small{
    \textbf{Correspondence:} \href{junhua.ding@unt.edu}{junhua.ding@unt.edu}
  }
        }

\begin{document}
\maketitle
\begin{abstract}
Evaluating text summarization has been a challenging task in natural language processing (NLP). Automatic metrics which heavily rely on reference summaries are not suitable in many situations, while human evaluation is time-consuming and labor-intensive. To bridge this gap, this paper proposes a novel method based on large language models (LLMs) for evaluating text summarization. We also conducts a comparative study on eight automatic metrics, human evaluation, and our proposed LLM-based method. Seven different types of state-of-the-art (SOTA) summarization models were evaluated. We perform extensive experiments and analysis on datasets with patent documents.  Our results show that LLMs evaluation aligns closely with human evaluation, while widely-used automatic metrics such as ROUGE-2, BERTScore, and SummaC do not and also lack consistency. Based on the empirical comparison, we propose a LLM-powered framework for automatically evaluating and improving text summarization, which is beneficial and could attract wide attention among the community.
\end{abstract}

\section{Introduction}

Text summarization is the process of producing a concise and coherent summary while preserving key information and meaning of the source text \cite{allahyari2017text}. This technique is widely used in various fields; for example, it is commonly used to summarize scientific, medical, and legal documents, as it enables users to quickly grasp key points of lengthy texts and efficiently access relevant information. 

There are two major approaches to automatic text summarization: extractive and abstractive. Extractive summarization involves selecting important sentences or phrases from the original document. 
Extractive summarization is considered to be faster, simpler and more accurate because it retains authentic sentences of the source documents. However, it is less fluent and less coherent than abstractive summarization \cite{el2021automatic}. On the other hand, the abstractive summary generates the summary with sentences that are different from those in the original text while not changing the central facts and ideas.  

With the remarkable achievements of pretrained language models (PLMs) and natural language generation, recent research has shifted gears from extractive to abstractive summarization. Nevertheless, the abstractive summarization still remains a challenging task since models suffer from hallucinations \cite{cao2018faithful, maynez2020faithfulness} and the generated summaries do not align with human expectations \cite{he2020ctrlsum}. A preliminary study of text summarization techniques indicates that nearly 30\% summaries generated by existing SOTA neural abstractive summarization are unfaithful to original documents \cite{cao2018faithful}. 

Summarization evaluation can be divided into reference-based and reference-free. Reference-based metrics are based on the matching between the generated summary and the reference summary, whereas the reference-free is based on the source document to evaluate the generated summary. Evaluation of summaries can be done either automatically or manually. The automatic method is fast, inexpensive, and can handle large volumes of data without human intervention. They can be categorized into three groups: text overlapping (e.g., ROUGE \cite{chin2004rouge,ganesan2018rouge} and BLEU \cite{papineni2002bleu}), vector-space distance (e.g., BERTScore \cite{zhang2019bertscore, kieuvongngam2020automatic}, MoverScore \cite{zhao-etal-2019-moverscore}), and NLP task-based to measure the consistency between the generated summary and the reference (e.g., SummaC \cite{laban2022summac}, QuestEval \cite{scialom2021questeval}). 

Summarization evaluation is still a challenging task as there are no ideal evaluation methods. Some studies demonstrate that automatic evaluation metrics such as BLEU, ROUGE, and BERTScore are not suitable for the automatic evaluation of summaries \cite{sun2022bertscore, sulem2018bleu, reiter2018structured, schluter2017limits}. Most of these automatic metrics, especially text overlapping and vector-space similarity measurements, are originally reference-based; therefore, they may not be appropriate to use as reference-free due to the incompatible length and information compression.

On the other hand,  the human evaluation is considered more reliable and trustworthy. It is still the preferred choice for evaluating summaries \cite{deutsch2021towards}. The process starts by sampling a small set of about 30-100 generated summaries. The recruited evaluators are asked to score the generated summary on a Likert scale such as 1-5, or 1-7 on one or more evaluation dimensions described above. Despite the advantages of human evaluation, conducting this evaluation method is time-consuming, and labor-intensive, so it is infeasible to use in model development. 

Studies that access the evaluation methods reach inconsistent conclusions. For example, \citet{graham2015re} claim that text-overlap metrics achieve the strongest correlation with human assessment. Meanwhile, recent studies demonstrate that metrics such as BLEU,
ROUGE, and BERTScore are not suitable for the automatic evaluation of summaries \cite{sun2022bertscore, sulem2018bleu, schluter2017limits, reiter2018structured}. Therefore, there is a need to revisit existing summarization evaluation methods.



In this work, we conduct a comparative study to reevaluate existing automatic metrics to evaluate abstractive summarization. Besides, we assess the ability of LLMs to perform as an evaluation agent. We further propose a framework to iteratively improve the quality of LLMs-generated summaries. Our contributions can be summarized as follows: 
\begin{itemize}
    
    \item We conduct a comprehensive evaluation of the latest off-the-shelf PLMs and LLMs for the patent document summarization, using both automatic and human evaluation methods.
    \item We re-evaluate existing automatic metrics that are widely-used for evaluating text summarization.
    \item We propose a framework based on LLMs for automatically evaluating and improving summarization.  
    
\end{itemize}

\section{Related Work}

\subsection{Text Summarization}

Most current abstractive models rely on neural networks based sequence-to-sequence learning \cite{bahdanau2015neural, vaswani2017attention}. Seq2seq summarization can be summarized into two main types of frameworks, RNN encoder-decoder \cite{bahdanau2015neural} and Transformer encoder-decoder \cite{vaswani2017attention}. 

\citet{nallapati2016abstractive} introduced one of the first RNN-based summarization models, utilizing a bidirectional RNN encoder enriched with POS tags and TFIDF feature embeddings, and a unidirectional RNN decoder with an attention mechanism. However, these models often faced issues with out-of-vocabulary (OOV) words and word repetition. To address these problems, \cite{see2017get} proposed a pointer-generator network, which combines the base seq2seq model with a pointer network that decides whether to generate a word from the vocabulary or copy it from the input sequence. Additionally, a coverage mechanism was implemented to track and prevent repetition \cite{see2017get, nallapati2016abstractive}. 

Abstractive summarization using Transformer encoder-decoder framework has rapidly advanced in recent years. Transformers with self-attention layers allow parallelization learning, solving the vanishing or explosion gradient of standard RNNs. It achieves SOTA performance in machine translation \cite{vaswani2017attention}. Given this success, this approach is promising in abstractive summarization. Currently, encoder-decoder Transformer models like BART \cite{lewis2020bart} and PEGASUS \cite{zhang2020pegasus} have achieved SOTA summarization results on short text. However, BART’s and PEGARUS's maximum input length limit at 1024 tokens, making it unsuitable for summarizing long text.

The major limitation of transformer models is the complexity of quadratic self-attention that grows rapidly with sequence length \cite{zaheer2020big}. This has significantly impeded their effectiveness in summarizing long documents. The simplest approach is truncating the document from the head or tail to produce a short valid input. However, \cite{meng2021bringing} proves that Transformers with this naive method is even worse than many unsupervised algorithms, such as TextRank, LSA, etc., for long text summarization. Models like Longformer \cite{beltagy2020longformer} and BigBird \cite{zaheer2020big} incorporate sparse attention mechanisms in the encoder to reduce the computational cost of standard self-attention operation \cite{vaswani2017attention}; therefore, it can handle longer contexts. Longformer which is a Transformer model that uses the BART architecture can take up to 16k input tokens while Bird can support a sequence length of 4k tokens. 
The proposed attention replaces the full self-attention in standard Transformers \cite{vaswani2017attention} with the attention pattern mechanism, including windowed, dilated, and global attention. The model achieved SOTA performance on arXiv dataset, a long text summarization dataset, surpassing BARD-base, Pegasus, and BigBird \cite{beltagy2020longformer}. 
BigBird reduces quadratic complexity with a sparse attention mechanism combining random, windowed, and global attentions, allowing it to handle longer sequences efficiently without significantly increasing computational resources.

The model pretraining is continued from Pegasus \cite{zhang2020pegasus} that is specified for abstractive summarization. The model performs better than base Transformers, Pegasus, BIGBIRD-RoBERTa, etc. on three long-text summarization datasets, including BigPatent, arXiv, and Pubmed  \cite{zaheer2020big}. 

\subsection{Summarization Evaluation}

Summarization systems can be evaluated with or without a reference summary. \textit{Reference-based} evaluation uses reference summaries to identify what content from the input document is important and then evaluates a generated summary based on how similar it is to the reference. On the other hand, \textit{reference-free} evaluation directly or indirectly defines a model to capture important information in the document and uses that to evaluate the content of the candidate summary. 
The recent success of LLMs has raised a lot of attention about how to evaluate the generated summaries since the reference summaries are too generic or unavailable. Therefore, in this study, we focus on assessing the \textit{reference-free} evaluation methods.

Evaluation of summaries can be done either automatically or manually. The automatic method is fast, inexpensive, and can handle large volumes of data without human intervention. However, this method may not be able to measure the exact aspects of summaries that humans are interested in evaluating such as clarity, accuracy, coverage, etc. On the other hand, manual evaluation is slower, more expensive, and infeasible to use in model development, but it is considered more reliable and trustworthy. 
A detailed discussion of the automatic and human evaluation method is presented in Appendix \ref{sec:exist-eval-method}. We also present an overview of summarization quality evaluation based on existing studies in  Figure \ref{fig:eval-ontology} in Appendix \ref{sec:exist-eval-method}.
\label{sec:sum_eval}

\section{Methodology}

\subsection{Summarization Models}

Summarization models studied in this research are SOTA PLMs,  including the T5 family, XLNet, BART, BigBird, Pegasus, and GPT-3.5. These models have demonstrated significant potential across various applications and are categorized as follows: (1) domain-specific models (HUPD\_T5\_small and HUPD\_T5\_base), (2) general-domain models (XLNet, BART, and Pegasus), (3) models for long input sequences (LongT5 and BigBird), and (4) large language models (LLMs) like GPT-3.5. This selection ensures a comprehensive evaluation covering diverse types of SOTA summarization models.

Text-To-Text Transfer Transformer (\textbf{T5 family}), developed by Google, is an encoder-decoder Transformer designed for a variety of NLP tasks\cite{raffel2020exploring} thanks to its capability to convert any language task into an essential text-to-text task. We implement the two off-the-shelf T5-based models:

\texttt {\textbf{HUPD-T5}} \cite{suzgun2022harvardpublished}
    Harvard University's Policy Department (HUPD) has tailored the T5 model for legal document summarization. We employ two versions of this model: {\texttt{hupd-t5-base}} and {\texttt{hudp-t5-small}}. The two model versions are finetuned for the abstractive summarization task on the Harvard USPTO Patent Dataset (HUPD) \cite{suzgun2024harvard}, a large-scale corpus of utility patent applications filed to the United States Patent and Trademark Office (USPTO) between January 2004 and December 2018. Therefore, these models are ideal for the legal document summarization task that we focus on.

\texttt{\textbf{long-t5-tglobal-base-16384 + BookSum}} (so-called \texttt{LongT5}) \cite{zhang2023summit}: The model is based on T5 with an expanded context window of 16384 tokens, enabling it to comprehend and summarize long text efficiently\cite{peter_szemraj_2022}. The model is trained on a large corpus of book summaries, providing it with a strong capability to digest and generate summaries of long texts \footnote{\url{https://huggingface.co/pszemraj/long-t5-tglobal-base-16384-book-summary}}. 

\textbf{XLNet \cite{yang2020xlnet}:}  an autoregressive transformer model, has 110 million parameters and is trained on an assortment of datasets such as BooksCorpus, Wikipedia, and Giga5\cite{yang2020xlnet}. Its distinguishing feature is the permutation-based training, allowing the model to capture bidirectional contexts and understand the intricate dependencies and relationships within the text.
    
\textbf{BART~\cite{lewis2020bart}:} is an encoder-decoder transformer model with 140 million parameters. The model was pre-trained on a wide variety of data sources such as BookCorpus, Wikipedia, news articles, and stories \cite{lewis2020bart}. The model is popular for text summarization and question-answering tasks.

\textbf{BigBird \cite{zaheer2020big}:} is a transformer model that pioneers the use of a sparse attention mechanism with 110 million parameters. It is trained on diverse data sources such as Wikipedia, BookCorpus, and news articles. The introduction of the sparse attention mechanism enables BigBird to efficiently manage very long sequences, thus, reducing the computational complexity that is generally associated with processing long-range dependencies in text~\cite{zaheer2020big}. It is suitable for processing lengthy texts due to its sparse attention mechanism. However, this design might capture less context compared to models with complete attention mechanisms, potentially affecting the overall quality of the model's outputs.

\textbf{Pegasus \cite{zhang2020pegasus}:} is an encoder-decoder transformer model with the pretraining objective specifically optimized for summarization. It is trained on diverse datasets like C4, HugeNews, PubMed, and arXiv. We use \texttt{pegasus-x-large-booksum-1}, an off-the-shelf model with 568 million parameters for the summarization experiments. 
    
\textbf{GPT-3.5 \cite{liu2023lost}: } Recent GPT models have demonstrated unprecedented capabilities in understanding context, semantics, and syntactic structures, enabling them to generate summaries that are concise, coherent, and human-like. Experiments on multiple news datasets, \citet{pu2023summarization} even found that humans significantly prefer summaries generated by zero-shot GPT-3.5 and GPT-4 to those written by humans or generated by small finetuned PLMs such as BART \cite{lewis2020bart}, T5 \cite{raffel2020exploring}, etc. In this study, we use \texttt{GPT-3.5-turbo-16k} for summarization. It features an extended context window handling up to 16k tokens,  enabling it to maintain context over longer conversations for more accurate and coherent responses. 

\subsection{Dataset}

The dataset for this study consists of a corpus of 1630 patent documents collected through web scraping of Google Patents \footnote{\url{https://patents.google.com}}.  Human evaluation requires a good understanding of the documents. To simplify the evaluation process, we have focused on collecting patents related to communication and streaming technologies. 

Although a patent document includes long description of the invention details and many flow charts, the most important content in a patent document for the summarization includes its abstract and the claims of the invention. The abstract provides an overview of the invention, while the claims detail the invention's specifics that a summarization must capture. While other datasets, such as BIGPATENT\cite{sharma2019bigpatent}, consider the abstract as an abstractive summary of the patent document, the abstract itself does not cover the scope of the claim and the novelty of the patent. To generate more useful and self-contained patent summaries, we use the abstract and the claims from each patent document as inputs for the summarization models.

\subsubsection{Data Sample for Evaluation}
Since human evaluation is only conducted in a small-scale, we randomly sample a subset of 30 patent documents to generate summaries for evaluation with humans. Figure \ref{fig:length-dis} in Appendix \ref{sec:apdx-dataset} shows the text-length distribution of the entire dataset and the evaluation sample.
Besides, due to the high cost of human evaluation, we only evaluate summaries from five summarization models, including HUPD\_T5\_base, XLNet, BART, LongT5, GPT-3.5, and Llama-3.  We select representative models from each group based on their performance on traditional automatic metrics. 

\subsection{Evaluation Methods}

Many metrics have been proposed for text summarization evaluation; however, not all of them are suitable for evaluating summarization in long text. We select eight widely-used automatic evaluation metrics for this study. In addition, we also conduct human evaluation and introduce a LLM-based evaluation approach. 

\subsubsection{Automatic Metrics}

\textbf{ROUGE-1:} measures the unigram overlap between the candidate summary (generated summary) and the reference summary (ground truth summary). It calculates the proportion of overlapping unigrams (individual words) between the candidate and reference summaries. It considers only individual words and does not capture word order or context~\cite{lin2004rouge}.

\textbf{ROUGE-2:} measures the bigram overlap between the candidate summary and the reference summary. It calculates the proportion of overlapping bigrams (consecutive pairs of words) between the candidate and reference summaries. It captures some level of word order and context by considering pairs of words together~\cite{lin2004rouge}.

\textbf{ROUGE-L:} measures the longest common subsequence between the candidate summary and the reference summary. It finds the longest sequence of words that appear in the same order in both the candidate and reference summaries. ROUGE-L accounts for word order and captures the informativeness of the candidate summary by considering sequences of words rather than just individual words~\cite{lin2004rouge}.

\textbf{Bilingual Evaluation Understudy (BLEU)} is a widely-used metric for evaluating the quality of machine-generated text, especially in machine translation. The BLEU score quantifies the similarity between the machine-generated text and one or more reference texts. It considers the matching n-grams between the generated text and the reference text. BLEU combines the scores for different n-grams (usually 1 to 4) into a single score by calculating the geometric mean of the modified precision counts \cite{papineni2002bleu}. The final BLEU score ranges from 0 to 1; 1 indicates a perfect match with the reference, while 0 indicates no overlap in n-grams. 

\textbf{BERTScore} is an evaluation metric for language generation based on pretrained BERT contextual embeddings \cite{devlin2018bert}. The metric computes a similarity score for each token pair between the generated text and the reference text using contextual embeddings. BERTScore has been shown to correlate well with human judgment of text quality \cite{zhang2019bertscore}. It has been used to evaluate a variety of text generation tasks, including machine translation, summarization, and question-answering. The scores, such as Precision, Recall, and F1, range from 0 to 1, with higher scores indicating better performance.

\textbf{SummaC} measures the consistency between a summary and its source text. The score is calculated by comparing the generated summary to the source text by identifying any inconsistencies between them.  SummaC effectively utilizes NLI models for inconsistency detection by segmenting documents into sentence units and aggregating scores between pairs of sentences \cite{laban2022summac}. The authors introduce two versions: $SummaC_{ZS}$, and $SummaC_{Conv}$ in which $SummaC_{ZS}$ use an out-of-the-box NLI model while $SummaC_{Conv}$ involves finetuning using a convolutional neural network. $SummaC_{Conv}$ achieves better performance on the proposed benchmark for the summary consistency detection task than $SummaC_{ZS}$ and other models such as QuestEval \cite{scialom2021questeval}, FactCC-CLS, etc. Therefore, we utilize $SummaC_{Conv}$ to evaluate the faithfulness of summarization models.

\textbf{Flesch Reading Ease (FRE) score} \cite{flesch1979write} assesses the readability of an English text by examining the sentence length and word length. It is calculated as $FRE = 206.835 - (1.015 * ASL) - (84.6 * ASW)$, where $ASL$ is the average sentence length and $ASW$ is the average number of syllables per word. The score typically ranges from 0 to 100. Higher scores indicate that the text is easier to read, while lower scores indicate that the text is more difficult to read. 

\textbf{Dale-Chall Readability (DCR) score} is another readability metric used to assess the readability of English text. It considers a set of familiar words and examines the sentence length to estimate the text's difficulty level. The DCR score is calculated: $DCR = (0.1579 * PDW * 100) + (0.0496 * ASL)$, where $PDW$ is the percentage of difficult words in the text, and $ASL$ is the average sentence length. DCR provides an estimated percentage value representing the difficulty level. Lower DCR scores indicate higher difficulty. DCR and FRE have drawbacks. They primarily considers sentence length and the presence of difficult words but does not account for factors like content, coherence, or text structure, which also influence readability.

\subsubsection{Human Evaluation}

Existing automatic evaluation metrics do not consistently align with human expectations, and there is no available human evaluation data for legal-text summarization. Therefore, we conducted a manual evaluation study with participants who are master’s students in computer and engineering fields. Given the technical nature of the texts, participants needed a solid understanding of such documents. To ensure quality, we designed test questions mixed with other questions to filter out unreliable responses from participants who failed the tests.

Based on our overview of existing summarization evaluation (Figure \ref{fig:eval-ontology}), we choose to evaluate the following quality dimensions: (1) \textit{Clarity}: whether the summary is reader-friendly and expresses ideas clearly; (2) \textit{Accuracy}: whether the summary contains the same information as the source document; (3) \textit{Coverage}: how well the summary covers the important information from the source document; and (4) \textit{Overall quality}: how good the summary overall at representing the source document; a good summary is a shorter piece of text that has the essence of the original – tries to convey the same information as the source document.  The participants are asked to rate the given summary on the above-mentioned quality dimensions on a Likert scale from 1 to 5, corresponding to Poor to Excellent. Our study uses APPEN platform \footnote{\url{https://client.appen.com/sessions/new}} to design and conduct the evaluation. Figure \ref{fig:screenshot_appen_interface} shows the interface of the data collection form with APPEN. We also release the response form of the study for the purpose of reproduction. \footnote{\href{https://account.appen.com/channels/cf\_internal/jobs/2377354/work?secret=2g7C\%2FLK\%2BeVn\%2FDg\%2FdFFjyjXnTJLz2LC7c8vBLPNOhG6zo}{Response form in APPEN}}


\subsubsection{LLM-Based Evaluation}
LLMs have demonstrated the exceptional ability to understand and follow instructions. They potentially can serve as an evaluation agent \cite{wu2023large, chiang2023can}. In this study, we explore the ability of LLMs to assess the quality of model-generated summaries. To ensure a fair comparison between LLMs and humans, we use the same instructions given to humans as prompts to guide LLMs on this task. Similarly, we ask LLMs to evaluate the summary quality on the four quality dimensions mentioned above on a similar Likert scale 1-5. 

\subsection{Summarization Improvement Based on LLM's Feedback}
Motivated by the way humans refine written text,  self-refining \cite{madaan2024self} and self-reflection \cite{shinn2024reflexion} have been proposed to enhance the initial text generated by LLMs. In the context of summarization, the initial draft iteratively refined will improve the quality. We adopt this approach for summarization improvement. Particularly, we iteratively incorporate the LLM's evaluation of the summary generated in the previous round into prompt to guide LLMs in generating a better version in the next round. We provide the prompt template that we use for this experiment in Appendix \ref{sec:apdx_improve_sum}.

\section{Results and Discussion}

\subsection{Evaluation on Summarization Performance}

\paragraph{Using existing automatic methods} To evaluate the performance of summarization, we use the most widely-used automatic evaluation metrics, including BLEU, ROUGE, BERTScore, $SummaC$, FRE, and DCR. While BLEU, ROUGE, BERTScore, and $SummaC$ are used to assess the content quality of generated summaries,  FRE and DCR are metrics to evaluate the readability level. Table \ref{tab:auto-metric-perf} shows performance of summarization models on these metrics on the entire dataset (left) and on evaluation sample (right). On the content-based evaluation metrics, the results show that XLNet, BART, and GPT-3.5 are the best models. However, their generated summaries are as not as readable as T5-generated summaries. BLEU metric yields low scores ($<<0.01$) for all summarization models, indicating a need to reconsider using this metric for evaluation. XLNet performs best on ROUGE and BERTScore, followed by GPT-3.5, and then BART. BigBird performs the worst among the eight models. The results conducted on the entire dataset are consistent with the results on the evaluation sample data that we sampled for human and LLM-based evaluation. 
\label{sec:results-auto-metric}

\begin{table*} [!ht]
\scriptsize
\begin{center} 

   \caption{Performances of summarization models on automatic evaluation metrics. Results are reported for the entire dataset and the evaluation sample of 30 source documents.}
   \label{tab:auto-metric-perf}
   \begin{threeparttable}
   \resizebox{\textwidth}{!}{
   \Huge
   \begin{tabular}{cccccccccc|ccccccccc}
   \hline
    & \multicolumn{9}{c|}{Entire Dataset} & \multicolumn{9}{c}{Evaluation Sample} \\
    \cline{2-19}
    & HTS & HTB & X & B & BB & P & LT5 & GPT & Ll & HTS & HTB & X & B & BB & P & LT5 & GPT & Ll \\
    \hline
    BLEU & 0.0001 & 0.0000 & \cellcolor{blond}0.0028 & 0.0700 & 0.0004 & 0.0000 & 0.0000 & \cellcolor{lightgreen}0.0046 & - & - & 0.0(0.0) & 0.0(0.0) & 0.0(0.0) & - & - & 0.0(0.0) & 0.002(0.01) & 0.00(0.00) \\
    Rouge-1 & 0.2851 & 0.2261 & \cellcolor{lightgreen}0.5298 & 0.3665 & 0.1142 & 0.3202 & 0.2727 & \cellcolor{blond}0.4239 & - & - & 0.233(0.05) & \cellcolor{lightgreen}0.523(0.07) & 0.366(0.06) & - & - & 0.297(0.05) & \cellcolor{blond}0.431(0.07) & 0.38(0.06 \\
    Rouge-2 & 0.1682 & 0.1187 & \cellcolor{lightgreen}0.3993 & \cellcolor{blond}0.2274 & 0.0308 & 0.0766 & 0.0471 & 0.2170 & - & - & 0.12(0.03) & \cellcolor{lightgreen}0.381(0.08) & \cellcolor{blond}0.221(0.05) & - & - & 0.052(0.02) & 0.208(0.07) & 0.18(0.06) \\
    Rouge-L & 0.2840 & 0.2249 & \cellcolor{lightgreen}0.5298 & 0.3656 & 0.1081 & 0.2906 & 0.2473 & \cellcolor{blond}0.4053 & - & - & 0.232(0.05) & \cellcolor{lightgreen}0.523(0.07) & 0.366(0.06) & - & - & 0.268(0.05) & \cellcolor{blond}0.412(0.06) & 0.36(0.06)\\
    BERTScore & 0.6599 & 0.5985 & \cellcolor{lightgreen}0.7500 & \cellcolor{blond}0.6741 & 0.4882 & 0.5943 & 0.5600 & 0.6700 & - & - & 0.605(0.05) & \cellcolor{lightgreen}0.738(0.06) & \cellcolor{blond}0.677(0.04) & - & - & 0.571(0.04) & \cellcolor{blond}0.67(0.04) & 0.65(0.03)\\
    $SummaC$ & 0.8608 & 0.9189 & \cellcolor{blond}0.9232 & \cellcolor{lightgreen}0.9268 & 0.3095 & 0.5191 & 0.4310 & 0.7797 & - & - & 0.914(0.09) & 0.925(0.04) & \cellcolor{lightgreen}0.936(0.07) & - & - & 0.474(0.14) & 0.76(0.16) & 0.63(0.20) \\
    FRE & 36.2081 & \cellcolor{lightgreen}41.5381 & 23.3092 & 35.6559 & 23.0606 & 37.2800 & \cellcolor{blond}38.5907 & 31.2010 & - & - & \cellcolor{lightgreen}40.743(14.49) & 23.422(11.97) & 35.064(11.07) & - & - & \cellcolor{blond}38.428(12.8) & 31.116(10.66) & 30.89(9.23)\\
    DCR & 9.0983 & \cellcolor{blond}10.6610 & 9.8782 & \cellcolor{lightgreen}10.6644 & 6.9065 & 10.2207 & 10.1642 & 9.5479 & - & - & \cellcolor{blond}10.719(1.85) & 10.188(0.92) & \cellcolor{lightgreen}10.941(0.96) & - & - & 10.053(0.98) & 9.717(0.91) & 10.16(0.77)\\
    \hline
    \end{tabular}}
    \begin{tablenotes}
    \scriptsize
        \item Note:
        $SummaC$ is the $SummaC_{conv}$, and all reported Rouge scores are the Rouge F1.  \colorbox{lightgreen}{Green} and \colorbox{blond}{Yellow} indicate the best and the second-best model \\performance on a certain metric, respectively. HTS, HTB, X and B denote HUPD\_T5\_small, HUPD\_T5\_base, XLNet and BART 
        BB, P, LT5, GPT and Ll denote \\BigBird, Pegasus, LongT5, GPT-3.5 and Llama-3.
    \end{tablenotes}
\end{threeparttable} 
\end{center}
\end{table*}

\paragraph{Human and LLM-based evaluation} 
The human and LLM-based evaluation is conducted regarding the four dimensions, including clarity, accuracy, coverage, and overall quality. Table \ref{tab:human-LLM-results} present results of human and GPT-4 evaluation for the five selected summarization models.
Overall, GPT-3.5 produces the best-quality summaries with the highest scores on all evaluation dimensions, followed by XLNet and BART. XLNet and BART have comparable accuracy and coverage quality scores.
T5-base and LongT5 are the worst regarding the clarity. We further examine the summaries producted by these low-performance models to propose improvement methods. Our analysis results are presented in Appendix \ref{sec:apdx-error-analysis}.
Surprisingly, the GPT-4 evaluation is highly consistent with humans in assessing the performance of each model, demonstrating the potential of this automatic method to replace expensive human evaluation.

\begin{table*}[!ht]
\footnotesize
    \centering
     \caption{Human and LLM (GPT-4) summarization evaluation}
    \label{tab:human-LLM-results}
    \resizebox{\textwidth}{!}{
  \Huge
  \begin{tabular}{ccccccc|cccccc}
  \hline
    
    & \multicolumn{6}{c|}{Human evaluation} & \multicolumn{6}{c}{LLM (GPT-4) evaluation} \\
    \cline{2-13}
    & HTB & X & B & LT5 & GPT & Ll & HTB & X & B & LT5 & GPT & Ll \\
    \hline
        Clarity & 2.183(0.48) & 2.6(0.55) & \cellcolor{blond}3.083(0.59) & 2.267(0.68) & \cellcolor{lightgreen} 4.55(0.27) & - & 3.32(0.76) & 3.2(0.66) & \cellcolor{blond} 3.8(0.48) & 2.167(0.53) & \cellcolor{lightgreen} 4.167(0.38) & 4.300( 0.47) \\
        Accuracy & 2.017(0.36) & \cellcolor{blond}2.883(0.55) & 2.783(0.73) & 2.083(0.51) & \cellcolor{lightgreen} 4.35(0.44) & - & 2.2(0.55) & \cellcolor{blond} 2.9(0.61) & \cellcolor{blond} 2.867(0.43) & 1.567(0.5) & \cellcolor{lightgreen} 3.9(0.48) & 3.667(0.55)\\
        Coverage & 1.8(0.45) & \cellcolor{blond}2.517(0.56) & \cellcolor{blond}2.517(0.56) & 2.133(0.61) & \cellcolor{lightgreen} 4.517(0.33) & - & 1.433(0.5) & \cellcolor{blond} 2.033(0.49) & \cellcolor{blond} 2.0(0.26) & 1.4(0.5) & \cellcolor{lightgreen} 3.567(0.57) & 3.333(0.55) \\ 
        Overall quality & 1.7(0.43) & 2.467(0.47) & \cellcolor{blond}2.6(0.62) & 2.0(0.66) & \cellcolor{lightgreen} 4.45(0.3) & - & 2.0(0.64) & 2.467(0.63) & \cellcolor{blond} 2.7(0.54) & 1.467(0.51) & \cellcolor{lightgreen} 3.967(0.41) & 3.733(0.43) \\
         \hline
    \end{tabular}}
     \begin{tablenotes}
     \scriptsize
    \item Note: 
    Reported results are averaged over all evaluation samples with standard deviations in brackets.  Human evaluation scores are reported on the scale of 1-5, \\respectively. HTB, X, B, LT5, GPT and Ll denote HUPD\_T5\_base, XLNet, BART, LongT5, GPT-4 and Llama-3.
    \end{tablenotes}
\end{table*}

\begin{table*}[!ht]
{\tiny
\centering
     \caption{Results of meta-analysis among automatic and human evaluation (Kendall Tau-b)}
    \label{tab:meta-analysis-auto-human}
    \resizebox{\textwidth}{!}{
    \begin{tabular} {llllllllllll} 
    \hline
     & accuracy & overall & coverage & clarity & R-1 & R-2 & R-L & BERTscore & FRE & DCR & SummaC \\
     \hline
    accuracy & 1.0*** & 0.8** & 0.95** & 0.8** & 0.8 & 0.4 & 0.8 & 0.4 & -0.8 & -0.4 & 0 \\
    overall & 0.8** & 1.0*** & 0.95*** & 1.0** & 0.6 & 0.2 & 0.6 & 0.2 & -0.6 & -0.2 & 0.2 \\
    coverage & 0.95** & 0.95*** & 1.0*** & 0.95** & 0.738 & 0.316 & 0.738 & 0.316 & -0.738 & -0.316 & 0.105 \\
    clarity & 0.8** & 1.0** & 0.95** & 1.0*** & 0.6 & 0.2 & 0.6 & 0.2 & -0.6 & -0.2 & 0.2 \\
    R-1 & 0.8 & 0.6 & 0.738 & 0.6 & 1.0*** & 0.6* & 1.0*** & 0.6* & -1.0** & -0.2 & 0.2 \\
    R-2 & 0.4 & 0.2 & 0.316 & 0.2 & 0.6* & 1.0*** & 0.6* & 1.0** & -0.6* & 0.2 & 0.6 \\
    R-L & 0.8 & 0.6 & 0.738 & 0.6 & 1.0*** & 0.6* & 1.0*** & 0.6* & -1.0** & -0.2 & 0.2 \\
    BERTscore & 0.4 & 0.2 & 0.316 & 0.2 & 0.6* & 1.0** & 0.6* & 1.0*** & -0.6* & 0.2 & 0.6 \\
    FRE & -0.8 & -0.6 & -0.738 & -0.6 & -1.0** & -0.6* & -1.0** & -0.6* & 1.0*** & 0.2 & -0.2 \\
    DCR & -0.4 & -0.2 & -0.316 & -0.2 & -0.2 & 0.2 & -0.2 & 0.2 & 0.2 & 1.0*** & 0.6 \\
    SummaC & 0 & 0.2 & 0.105 & 0.2 & 0.2 & 0.6 & 0.2 & 0.6 & -0.2 & 0.6 & 1.0*** \\
    \hline
    \end{tabular}
    }
    \begin{tablenotes}
    \scriptsize 
    \item Note:
       *, **, *** for $p$-value $<$ 0.05, 0.01, 0.001, respectively. 
       R-1, R-2, and R-L denote Rouge-1, Rouge-2, and Rouge-L.
    \end{tablenotes}
}
\end{table*}

    

\subsection{Meta-analysis}
Meta-analysis explores the pairwise statistic correlation between evaluation methods. Automatic evaluation metrics should strongly correlate with human judgments. Since the automatic evaluation metrics we utilize have not been specifically assessed for legal document summarization. Therefore, we would like to re-evaluate these automatic evaluation metrics.

\subsubsection{Automatic Metrics vs. Human Evaluation}
Table \ref{tab:meta-analysis-auto-human} presents results of meta-analysis between the conventional automatic metrics and human evaluation. Results indicate that ROUGE-1 and ROUGE-L correlate with human evaluation on four dimensions (0.6-0.8 on Kendall Tau-b). However, the significance test shows that the correlation is not statistically significant. The scores between BERTscore and ROUGE-2 and  human evaluation show a low level of correlation (0.2-0.4). Surprisingly, SummaC, a factual consistency measurement, displays very weak or non significant correlation with human evaluation, and many other metrics such as ROUGE-1, ROUGE-L. Both readability metrics have negative correlation with most metrics, including human evaluation, ROUGE scores, and BERTscore. This implies that texts that are easier to read tend to score lower on the content-based metrics, possibly suggesting a trade-off betweeen complexity and ease-of-reading.


\subsubsection{LLMs vs. Human Evaluation} 
Table \ref{tab:gpt4-llamavshuman}  presents the results of a meta-analysis comparing evaluations by LLMs and humans.  The correlation scores for the three quality dimensions (accuracy, coverage, and overall) between humans and LLMs (GPT-4 and Llama-3-8B) show an extremely high positive correlation (0.8-0.9). Significance tests indicate that these correlations are statistically significant. For the clarity dimension, LLMs and human evaluation still exhibit a high positive correlation (0.67-0.8); however, the statistical tests reveal that the correlations are not statistically significant, suggesting that the observed relationship could be due to random variation. The correlation scores are consistent for both GPT-4 and Llama-3-8B. The results indicate (1) LLMs are capable of performing summarization evaluation as effectively as humans, and (2) open-sourced LLMs such as Llama-3-8B can produce reliable evaluations similar to GPT-4. In specific domains like legal documents, where the cost of human evaluation is prohibitively high, LLMs, including open-sourced models, could be utilized to assess summarization.

In a previous analysis, we only used one type of correlation coefficient for our meta-analysis (Spearman’s or Kendall’s Tau-b). Therefore, we further compare the correlation results between GPT-4 and human evaluation using three tests: Pearson’s $\rho$, Spearman’s, and Kendall’s Tau-b correlation coefficients. The results indicate the consistency of these three correlation coefficients. Kendall's Tau-b produces lower scores than the two other tests. See Table \ref{tab:human-llm-3test}, Appendix \ref{sec:meta-3tests} for more details.

\begin{table*}[!ht]
{\tiny
\centering
    \caption{Results of meta-analysis between LLM (GPT-4) vs human evaluation and LLM (Llama-3-8B) vs human evaluation (Spearman's)}
    \label{tab:gpt4-llamavshuman}
    \begin{threeparttable}
    \resizebox{\textwidth}{!}{
    \Huge
    \begin{tabular}{lllllllll|llllllll}
    \hline
    & \multicolumn{8}{c|}{LLM (GPT-4) vs. human evaluation} & \multicolumn{8}{c}{LLM (Llama-3-8B) vs. human evaluation} \\
    \cline{2-17}
    & llm\_a & llm\_o & llm\_co & llm\_cl & h\_a & h\_o & h\_co & h\_cl & llm\_a & llm\_o & llm\_co & llm\_cl & h\_a & h\_o & h\_co & h\_cl \\
    \hline
    
    llm\_accuracy & 1.0*** & 0.9** & 1.0* & 0.821* & 0.9* & 0.8* & 0.872 & 0.8* & 1.0*** & 1.0** & 0.9** & 0.894 & 0.8* & 0.9* & 0.872* & 0.9* \\
    llm\_overall & 0.9** & 1.0*** & 0.9** & 0.975* & 0.8** & 0.9* & 0.872* & 0.9* & 1.0** & 1.0*** & 0.9* & 0.894 & 0.8* & 0.9 & 0.872 & 0.9* \\
    llm\_coverage & 1.0* & 0.9** & 1.0*** & 0.821 & 0.9*** & 0.8*** & 0.872** & 0.8** & 0.9** & 0.9* & 1.0*** & 0.783 & 0.9*** & 1.0** & 0.975** & 1.0** \\ 
    llm\_clarity & 0.821* & 0.975* & 0.821 & 1.0*** & 0.667 & 0.821 & 0.763 & 0.821 & 0.894 & 0.894 & 0.783 & 1.0*** & 0.783 & 0.783 & 0.803 & 0.783 \\
    human\_accuracy & 0.9* & 0.8** & 0.9*** & 0.667 & 1.0*** & 0.9** & 0.975** & 0.9** & 0.8* & 0.8* & 0.9*** & 0.783 & 1.0*** & 0.9** & 0.975** & 0.9** \\ 
    human\_overall & 0.8* & 0.9* & 0.8*** & 0.821 & 0.9** & 1.0*** & 0.975*** & 1.0** & 0.9* & 0.9 & 1.0** & 0.783 & 0.9** & 1.0*** & 0.975*** & 1.0** \\
    human\_coverage & 0.872 & 0.872* & 0.872** & 0.763 & 0.975** & 0.975*** & 1.0*** & 0.975** & 0.872* & 0.872 & 0.975** & 0.803 & 0.975** & 0.975*** & 1.0*** & 0.975** \\ 
    human\_clarity & 0.8* & 0.9* & 0.8** & 0.821 & 0.9** & 1.0** & 0.975** & 1.0*** & 0.9* & 0.9* & 1.0** & 0.783 & 0.9** & 1.0** & 0.975** & 1.0*** \\
    \hline
    \end{tabular}
    }
    \begin{tablenotes}
    \scriptsize 
    \item Note: *, **, *** for $p$-value $<$ 0.05, 0.01, 0.001, respectively.
     llm\_a, llm\_o, llm\_co, and llm\_cl denote llm\_accuracy, llm\_overall, llm\_coverage, and llm\_clarity. \\
    h\_a, h\_o, h\_co and h\_cl denote human\_accuracy, human\_overall, human\_coverage and human\_clarity. 
    \end{tablenotes}
    \end{threeparttable}
}
\end{table*}

\subsection{Summarization Quality Improvement}
Table \ref{tab:sum_improved} presents the results of our method for improving summarization It shows the impact of LLM's verbal feedback on summarization performance improvement. It demonstrates that by integrating evaluation feedback from LLMs into the prompt, the quality of the summaries significantly improves in terms of clarity (from 4.167 to 4.5) and coverage (from 3.567 to 3.833), indicating a substantial enhancement. However, we have also noticed that this method slightly reduces accuracy. In the future, we aim to enhance this quality dimension further.

\begin{table}[!ht]
{\footnotesize
    \centering
    \caption{Results of improving summarization (GPT-3.5) with LLM's verbal feedback. }
    \label{tab:sum_improved}
    \begin{tabular}{lll}
    \hline
     & \footnotesize w/o LLM's feedback & \footnotesize w/ LLM's feedback \\ \hline
    Clarity & 4.167(0.38) & 4.5(0.51) \textcolor{green}{$\uparrow$$\uparrow$$\uparrow$}\\ 
    Accuracy & 3.9(0.48) & 3.833(0.38) \textcolor{orange}{$\downarrow$}\\ 
    Coverage & 3.567(0.57) & 3.833(0.70) \textcolor{green}{$\uparrow$$\uparrow$} \\ 
    Overall quality & 3.967(0.41) & 3.933(0.41) \textcolor{orange}{$\downarrow$} \\ \hline
    \end{tabular}
    \begin{tablenotes}
        \scriptsize 
        \item Note: The arrows in the table indicate the direction of change: double arrows for \\significant improvements or declines, and single arrows for moderate changes.
    \end{tablenotes}
}
\end{table}


\section{Conclusion and Future Work}
In this work, we compare various methods for evaluating abstractive summarization of legal documents, including existing automatic metrics and human evaluation. We also explore the potential use of LLMs for evaluation purposes. We conduct different meta-analyses to compare the evaluations among these methods. Our findings reveal that widely-used automatic evaluation metrics such as ROUGE-2, BERTScore, and SummaC exhibit very weak or non-significant correlation with human evaluation. Additionally, readability metrics show a negative correlation with most other metrics, including human evaluation. In contrast, our results suggest that LLMs can effectively perform summarization evaluation. Open-sourced LLMs like Llama-3-8B demonstrates reliable evaluations similar to GPT-4.  Futher, we attempt to improve the summarization quality through iterative improvement by leveraging the verbal evaluation feedback from LLMs. The results indicate that the quality of the summaries significantly improves in terms of clarity and coverage, suggesting the potential of this approach. In the future, we aim to further enhance other quality dimensions such as accuracy.

\section{Limitations}
In this study, our focus is solely on evaluating the summarization of legal documents due to the high cost of human evaluation. Therefore, our findings may not be applicable to other domains where the document structure and vocabulary may differ. Additionally, the size of our human evaluation sample is limited, which may not accurately reflect the overall performance or reliability of the findings. This limitation could lead to the statistical insignificance of some meta-analyses we dicussed before and potentially cannot capture all possible variation within the dataset.

\newcommand{\ctext}[3][RGB]{%
  \begingroup
  \definecolor{hlcolor}{#1}{#2}\sethlcolor{hlcolor}%
  \hl{#3}%
  \endgroup
}


\section*{Acknowledgments}
We would like to thank Sai Kolapudi for his valuable help with data collection and preprocessing.

\bibliography{custom}
\appendix

\section{Appendix: Existing studies on text summarization evaluation} \label{sec:exist-eval-method}


\begin{figure*}[!ht]
    \centering
    \includegraphics[width=0.9\linewidth]{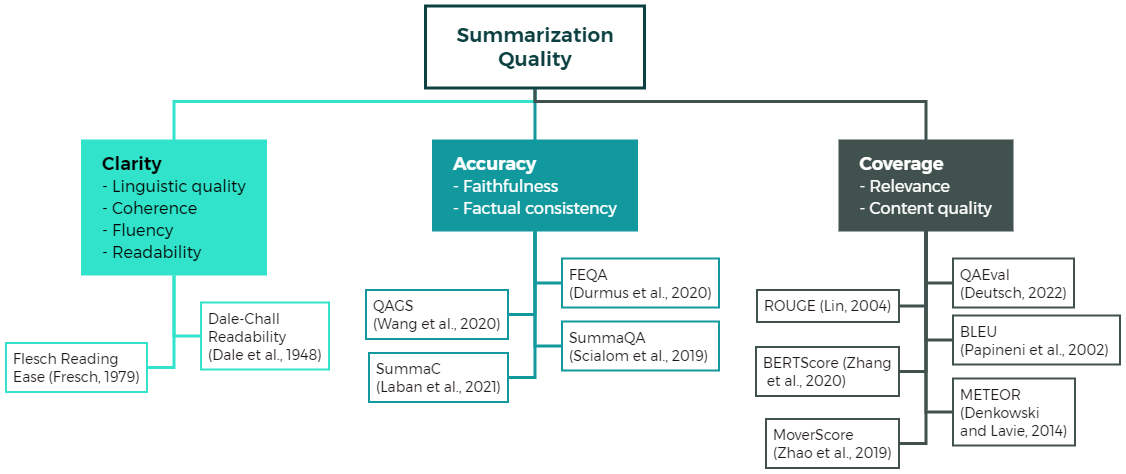}
    \caption{An Overview of Summarization Evaluation Methods}
    \label{fig:eval-ontology}
\end{figure*}

\subsection{Automatic methods}

Many studies have been done to determine which automatic metrics are sufficient for evaluation. 
They include ROUGE \cite{chin2004rouge,ganesan2018rouge}, BLEU \cite{papineni2002bleu}, BERTScore \cite{zhang2019bertscore, kieuvongngam2020automatic}, MoverScore \cite{zhao-etal-2019-moverscore}, SummaC \cite{laban2022summac}, QuestEval \cite{scialom2021questeval}, etc. They can be categorized into three groups: \textit{text overlapping} (including ROUGE and BLEU), \textit{vector-space distance }(BERTScore, MoverScore), and \textit{NLP task-based} to measure the consistency between the generated summary and the reference (SummaC, QuestEval).  


For the simplicity of the overlapping-based evaluation methods like ROUGE and BLEU, they are the most favorable to use. The use of text-overlapping metrics to assess the quality of summarization is still a topic of debate. 
\citet{graham2015re} uses summary coverage computations and human coverage scores to assert that text overlap-based metrics are suitable for evaluation. However, some studies demonstrate that evaluation tools such as BLEU, ROUGE, and BERTScore are not suitable for the automatic evaluation of summaries \cite{sun2022bertscore, sulem2018bleu, reiter2018structured, schluter2017limits}. Similarly, \citet{schluter2017limits} shows ROUGE's limitations, specifically, its inability to attain a perfect evaluation score. Therefore, most studies combine different metrics, such as text-overlap-based (ROUGE, BLEU), vector-space-based (BERTscore, MoverScore), and QA-based (QuestEval, SummaQA) to evaluate summarization performance.

On the other hand, some reference-free metrics have recently been proposed since summary references are not always available and high-quality, for example, 
QuestEval \cite{scialom2021questeval},  QAEval \cite{deutsch2021towards}. 
Compared to reference-free methods, reference-based metrics have more advantages. 
Additional studies also assert that reference-based metrics such as BERTScore correlate more closely with human judgments \cite{zhang2019bertscore}.

\subsection{Manual methods}
Human evaluation is commonly used because automatic metrics are imperfect, and humans can perform tasks that automated methods cannot do as reliably. The method involves asking human judges to score summaries based on the given reference (called \textit{reference-based}) or directly assess the generated summary according to specific criteria (called \textit{reference-free} or \textit{direct assessment}).

\paragraph{Reference-based} The manual reference-based approach is found to be simpler to conduct than the manual reference-free approach, which requires a well-designed scoring scheme and rubrics, and participants must have a good understanding of the source document. 

As mentioned earlier, we aim to explore the reference-free human evaluation methods in this study.

\paragraph{Reference-free} 

Most human evaluation methods for summarization are reference-free. Current works consider multiple dimensions to evaluate summary quality, including \textit{readability} (ease of reading), \textit{fluency} (grammaticality), \textit{consistency} (factual support from the input document), \textit{faithfulness} (completeness of information from the input document), \textit{relevance} (selection of important content), and \textit{content quality} (inclusion of salient information). Due to the high cost, evaluations often focus on fluency, coherence, consistency, and relevance.
The process starts by sampling a small set of about 50-100 generated summaries. The recruited evaluators are asked to score the generated summary on a Likert scale such as 1-5, or 1-7 on one or more evaluation dimensions described above. The results are then compared with automatic metrics using several correlation analysis tests such as Pearson's, Spearman's, and Kendall's Rank correlation coefficient.

Manual evaluation has been an important tool for measuring the quality of generated summaries. By focusing on specific dimensions and using a carefully selected group of evaluators, manual evaluation can provide valuable insights into the strengths and weaknesses of a summarization system.

\section{Appendix: Description of the APPEN platform used for evaluation} \label{sec:apdx-eval-platform}

\begin{figure*}
    \includegraphics[width=0.95\linewidth]{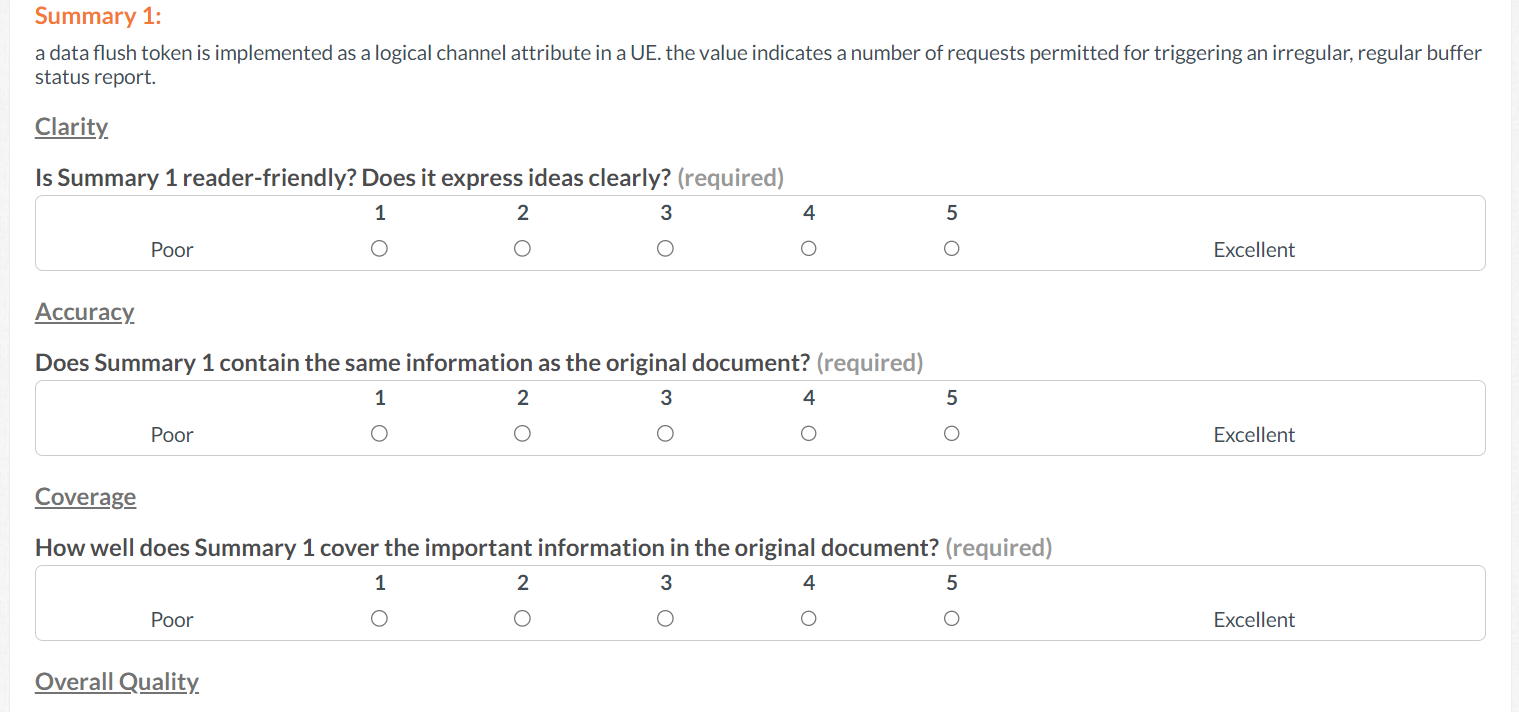}
    \caption{An example of the evaluation data collection form designed on APPEN platform. }
    \label{fig:screenshot_appen_interface}
\end{figure*} 

The APPEN platform was utilized for the manual evaluation of generated summaries in our study. APPEN provides a user-friendly interface that allows researchers to design and conduct surveys or evaluations efficiently. For this study, we created a form where participants could assess the quality of summaries based on predefined dimensions: Clarity, Accuracy, Coverage, and Overall Quality. Participants rated each summary on a Likert scale from 1 (Poor) to 5 (Excellent). The platform's robust functionality ensured accurate data collection and facilitated the identification of unreliable responses through carefully designed test questions. This systematic approach helped in obtaining reliable and consistent evaluations, which are crucial for assessing the performance of summarization models.

Figure \ref{fig:screenshot_appen_interface} illustrates the interface of the data collection form designed on the APPEN platform used in our manual evaluation study. Participants are presented with a summary and are asked to rate it across multiple dimensions: Clarity, Accuracy, Coverage, and Overall Quality. Each dimension is evaluated on a Likert scale from 1 to 5, ranging from Poor to Excellent. This interface ensures a structured and consistent approach to gathering evaluations from participants, facilitating the analysis of summarization quality against predefined criteria.

\section{Appendix: More detailed analysis of the datasets} \label{sec:apdx-dataset}

\begin{figure}[!ht]
    \centering
    \begin{subfigure}{0.5\textwidth}
        \centering
        \includegraphics[width=\linewidth]{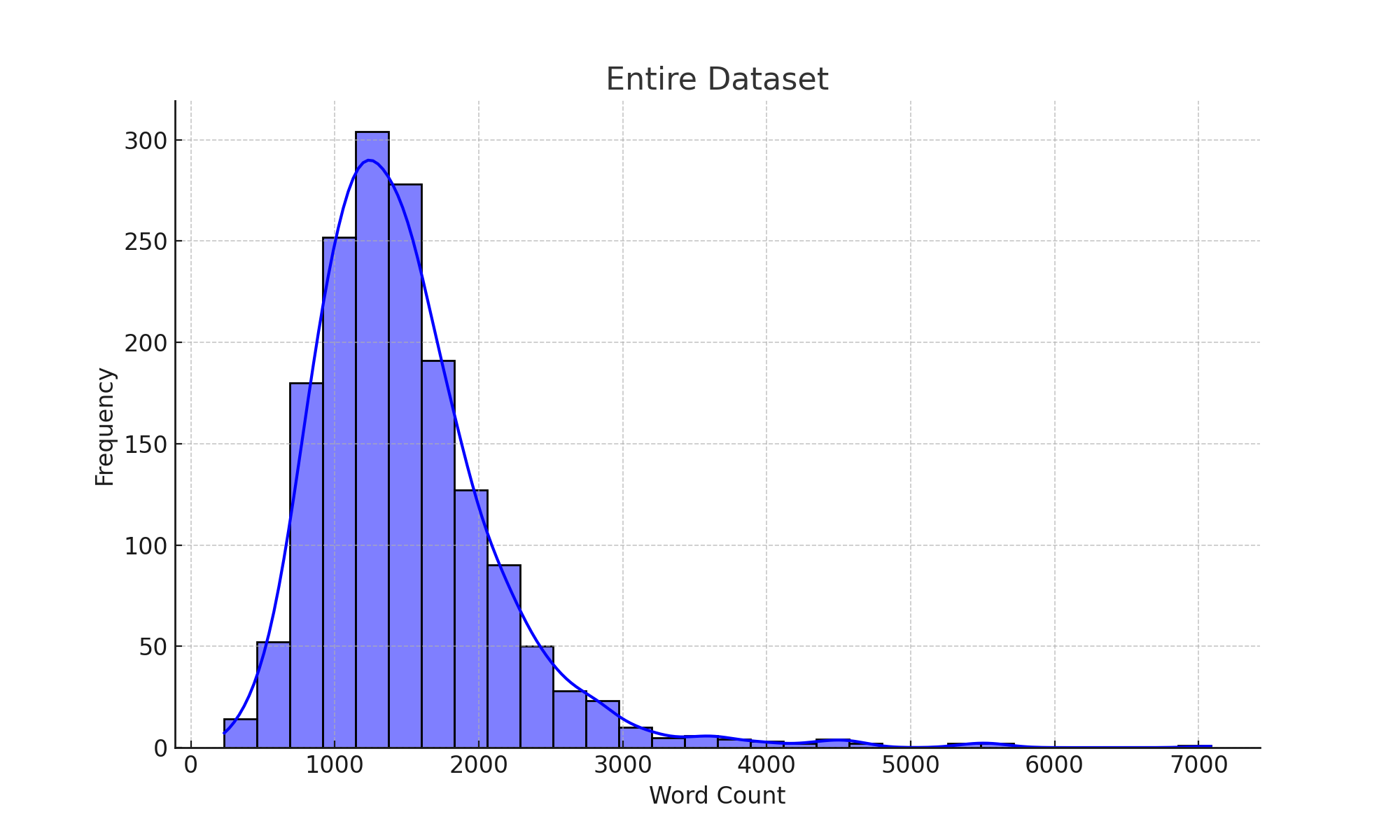}
    \end{subfigure}%
    \vspace{0.5cm}
    \begin{subfigure}{0.5\textwidth}
        \centering
        \includegraphics[width=\linewidth]{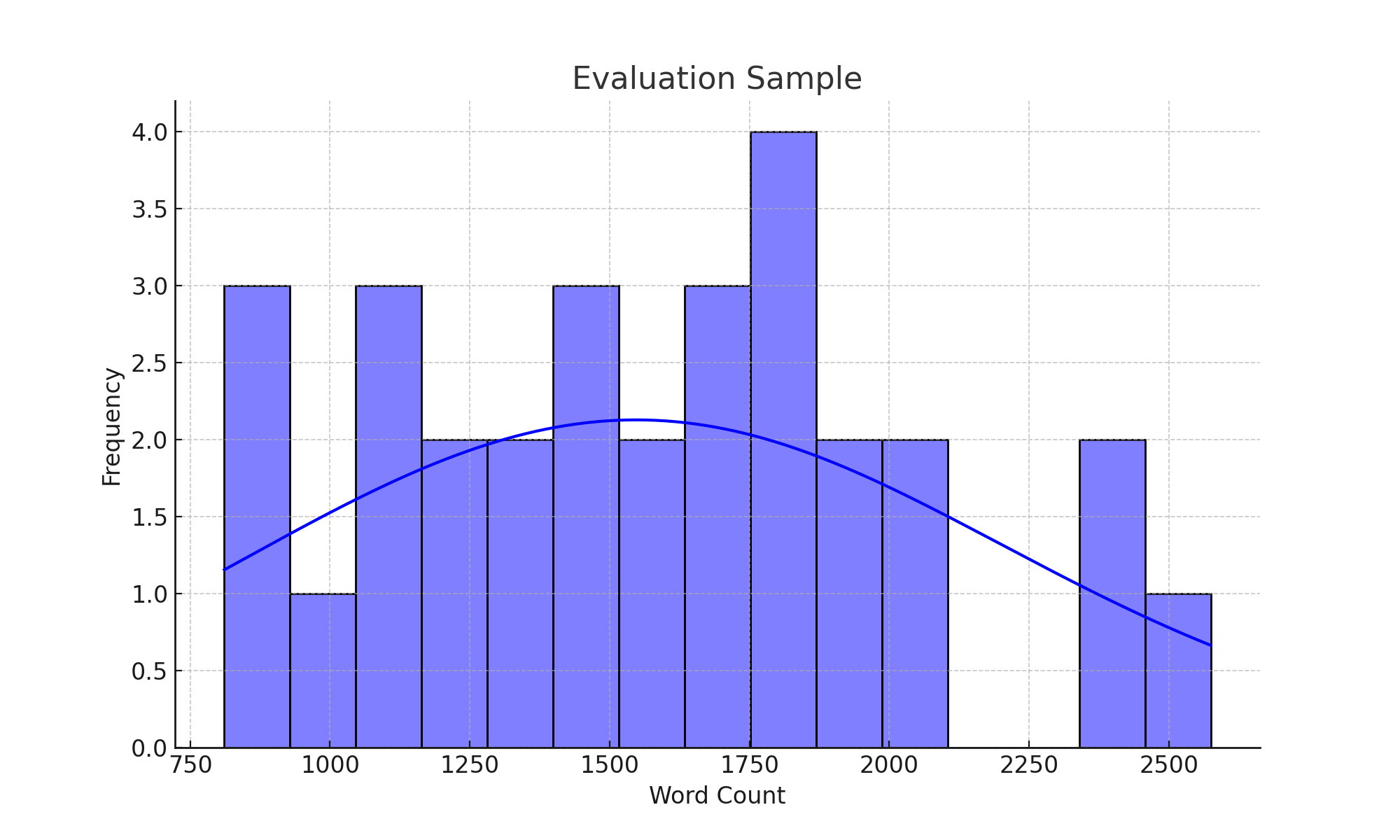}
    \end{subfigure}
    \caption{Length distributions of the entire dataset (top) and the evaluation sample (bottom).}
    \label{fig:length-dis}
\end{figure}

Figure \ref{fig:length-dis} contains two visualizations that shows the word count distribution for patent documents related to communication and streaming technologies, collected through web scraping of Google Patents. The first visualization presents the word count distribution for the entire dataset, consisting of 1,630 patent documents. The histogram displays the frequency of documents against their respective word counts, with the majority falling between 1,000 and 2,500 words, peaking around 1,500 words. This distribution indicates a right-skewed pattern with a long tail extending towards higher word counts, highlighting the typical length of patent documents used in the study, which focuses on abstracts and claims to generate comprehensive and informative summaries.

The second visualization shows the word count distribution for a randomly selected subset of 30 patent documents used for human evaluation. This subset was chosen from the larger dataset to ensure a diverse range of document lengths for evaluating the performance of various summarization models. The histogram displays the frequency of documents against their respective word counts, showing a relatively even distribution across different ranges, with most documents falling between 750 and 2,500 words and peaking around 1,750 words. 

\section{Appendix: Choosing a meta-analysis correlation coefficients}
\label{sec:meta-3tests}
Pearson’s $\rho$, Spearman’s, and Kendall’s Tau-b are widely-used correlation coefficients. We further conduct an experiment with these three coefficients to measure the strength of the relationship between GPT-4 and human evaluation scores. Table \ref{tab:human-llm-3test} present the result of this experiment. The results indicate the consistency of these three correlation coefficients. However, compared to two others, Kendall’s Tau-b produces lower scores. Therefore, when the correlation is not very strong, it may be harder to interpret the correlation results of Kendall’s Tau-b. In such cases, Pearson’s $\rho$ and Spearman’s coefficients may be more appropriate.

\begin{table*}[!ht]
\centering
\caption{Results of meta-analysis among LLM (GPT-4) and human evaluation using three different tests.}
\label{tab:human-llm-3test}
\resizebox{\textwidth}{!}{%
\begin{tabular}{ccllcllcllcll}
\hline
 & \multicolumn{3}{c}{human\_accuracy} & \multicolumn{3}{c}{human\_coverage} & \multicolumn{3}{c}{human\_clarity} & \multicolumn{3}{c}{human\_overall} \\
 \cline{2-12}
 & $r$ & \multicolumn{1}{c}{$\rho$} & \multicolumn{1}{c}{$\tau$} & $r$ & \multicolumn{1}{c}{$\rho$} & \multicolumn{1}{c}{$\tau$} & p & \multicolumn{1}{c}{$\rho$} & \multicolumn{1}{c}{$\tau$} & $r$ & \multicolumn{1}{c}{$\rho$} & \multicolumn{1}{c}{$\tau$} \\
\hline
llm\_accuracy & 0.939* & 0.9* & 0.8* & 0.866 & 0.872 & 0.738 & 0.896* & 0.8* & 0.6* & 0.889* & 0.8* & 0.6* \\
llm\_coverage & 0.996*** & 0.9*** & 0.8*** & 0.989** & 0.872** & 0.738** & 0.981** & 0.8** & 0.6** & 0.992*** & 0.8*** & 0.6*** \\
llm\_clarity & 0.777 & 0.667 & 0.527 & 0.694 & 0.763 & 0.667 & 0.795 & 0.821 & 0.738 & 0.732 & 0.821 & 0.738 \\
llm\_overall & 0.965** & 0.8** & 0.6** & 0.924* & 0.872* & 0.738* & 0.958* & 0.9* & 0.8* & 0.942* & 0.9* & 0.8* \\
\hline
\end{tabular}%
}
\begin{tablenotes}
    \scriptsize 
    \item Note:
       *, **, *** for $p$-value $<$ 0.05, 0.01, 0.001. $r$, $\rho$, and $\tau$ denote Pearson's $\rho$, Spearman's, and Kendall's Tau-b correlation coefficients, respectively.
    \end{tablenotes}
\end{table*}

\section{Appendix: Improving Summarization Quality}
\label{sec:apdx_improve_sum}
\begin{figure} [!ht]
    \centering
    \includegraphics[width=1\linewidth]{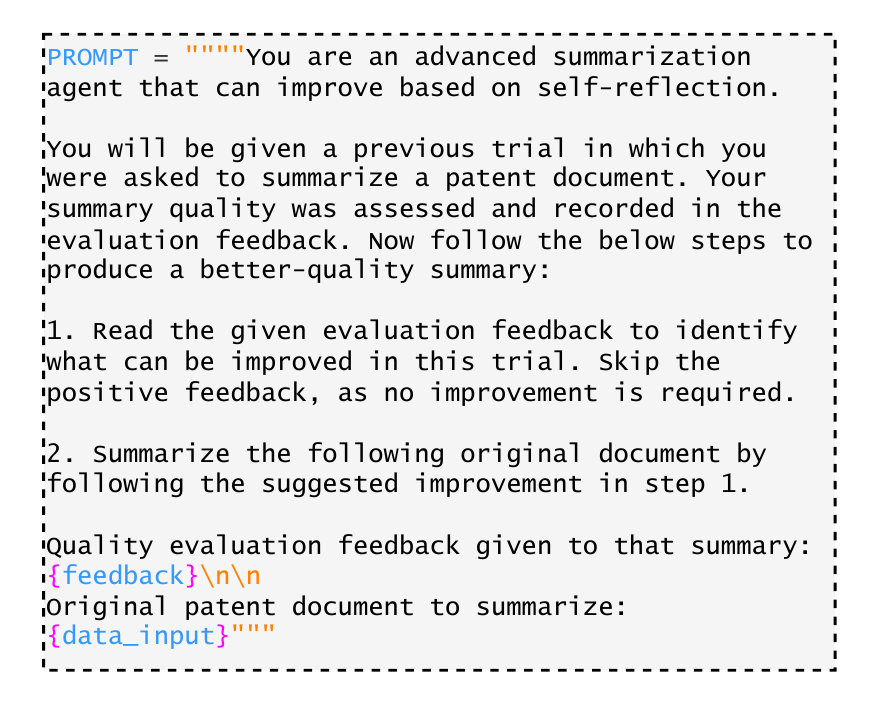}
    \caption{Prompt designed for iteratively improving summarization quality}
    \label{fig:improving_sum_prompt}
\end{figure}

\section{Appendix: Summarization models' error analysis} \label{sec:apdx-error-analysis} 
We conducted an error analysis to identify the specific areas where the summarization models underperformed. The purpose was to better understand the types and root sources of the errors and to provide guidelines for model selection and improve the model. We identified three major types of errors that affect the quality of summarization: low abstractiveness, incompleteness, and hallucinations. The low abstractiveness issue happens the most in the summaries generated by HUPD\_T5\_base, XLNet, and BART. 

Table \ref{tab:errors-exp} presents some selected examples of error types for illustration. The examples of HUPD\_T5\_base and XLNet generated summaries show that the models mostly copied long phrases or even entire sentences from the abstract and claims instead of generating new summary sentences. This likely led to the high performance of these models on ROUGE metrics which are based on n-gram overlaps.
HUPD\_T5\_base often copies and then combines long phrases to create a summary. However, its summaries do not have complete meaning, which makes them hard to understand. This is why the clarity score given to this model is the lowest, according to both humans and LLM evaluators (Table \ref{tab:auto-metric-perf}).
Furthermore, we found LongT5 suffered the most from the hallucination issue. The following example illustrates a significant hallucination by LongT5. In the source document, ``RIBS'' refers to ``radio interface based synchronization''. However, the model failed to recognize this and instead attempted to replace the term with ``bribs'', which was nonsensical in the context of the sentence.

\begin{table*}[!ht]
{\footnotesize
\centering
\caption{Examples of error types in generated summaries.}
\label{tab:errors-exp}

\begin{tabular}{|>{\centering\arraybackslash}p{1.7cm}|>{\raggedright\arraybackslash}p{6.7cm}|>{\raggedright\arraybackslash}p{5.5cm}|>{\centering\arraybackslash}p{2cm}|}
\hline
\textbf{Error type} & \textbf{Source document} & \textbf{Generated summary} & \textbf{Model}\\ 
\hline
Low abstractiveness & \textbf{Abstract:} The invention relates to a method, \ctext[RGB]{253,253,150}{comprising tracking the gaze of an eye of a user by a first camera}, wherein a position of the gaze determines a place for the cursor on the screen of a near-to-eye display (NED), wherein the movement of the gaze is configured to move the cursor on the screen and when the movement of the gaze is detected to be still or moving only slightly; observing a gesture of one hand of the user by a second camera, and executing a further operation if at least one gesture of the hand is detected. The invention further relates to an apparatus and a computer program product configured to perform the method. 
\textbf{Claims:}
 ... 
3. \ctext[RGB]{255, 179, 71}{The method of claim 1, wherein the detected gesture is a movement of a fist, shaking of a palm or movement of a palm from or to the second camera} ... &   \ctext[RGB]{255, 179, 71}{the method of claim 1} \ctext[RGB]{253,253,150}{comprises tracking the gaze of an eye by a first camera}. a second camera observes a gesture of one hand of the user. \ctext[RGB]{255, 179, 71}{the detected gesture} may be the \ctext[RGB]{255, 179, 71}{movement of a fist, shaking of a palm or movement of a palm}. & HUPD\_T5\_base \\
\hline
Low abstractiveness & \textbf{Abstract:}
        \ctext[RGB]{253,253,150}{Achieving phase synchronization in single frequency networks may be useful in various communication systems.} For example, the use of synchronous Ethernet, timing over packet, and radio interface based synchronization (RIBS) may be applied in mobile and/or data communication networks. Certain embodiments improve phase accuracy using a hybrid synchronization method beyond standard RIBS and clustering carriers using the same transmission waveforms across carriers. \ctext[RGB]{253,253,150}{Certain embodiments also select a transmission point to maximize the signal quality of the composite waveforms in a cluster.}
        \textbf{Claims:}
        \ctext[RGB]{255, 179, 71}{9.10. ...} & \ctext[RGB]{253,253,150}{Achieving phase synchronization in single frequency networks may be useful in various communication systems. Certain embodiments also select a master transmission point to maximize the signal quality of the composite waveforms in a cluster.} The apparatus according to claim 9, wherein the one or more quality metrics comprise an indication of at least one of signal power, signal to interference and noise ratio, path loss, and channel rank. The apparatus according to claim 9, wherein the at least one memory and the computer program code are further configured to, with the at least one processor, cause the apparatus to at least: receive, by the transmission point, phase synchronization data... &  XLNet \\
\hline
Incompleteness and potential hallucinations & \textbf{Abstract:}
        Achieving phase synchronization in single frequency networks may be useful in various communication systems. For example, the \ctext[RGB]{253,253,150}{use of synchronous Ethernet, timing over packet, and radio interface based synchronization (RIBS) may be applied} in mobile and/or data communication networks. \ctext[RGB]{253,253,150}{Certain embodiments improve phase accuracy using a hybrid synchronization method} beyond standard RIBS and clustering carriers using the same transmission waveforms across carriers. Certain embodiments also select a master transmission point to maximize the signal quality of the composite waveforms in a cluster.
        \textbf{Claims:} ... & \ctext[RGB]{253,253,150}{use of synchronous Ethernet, timing over packet, and radio interface based synchronization (RIBS) may be applied. certain embodiments improve phase accuracy using a hybrid synchronization method.} &  HUPD\_T5\_base \\
\hline
Hallucination &  \textbf{Abstract:}
        Achieving phase synchronization in single frequency networks may be useful in various communication systems. For example, the use of synchronous Ethernet, timing over packet, and \ctext[RGB]{255, 179, 71}{radio interface based synchronization (RIBS)} may be applied in mobile and/or data communication networks. Certain embodiments improve phase accuracy using a hybrid synchronization method beyond \ctext[RGB]{255, 179, 71}{standard RIBS} and clustering carriers using the same transmission waveforms across carriers. Certain embodiments also select a master transmission point to maximize the signal quality of the composite waveforms in a cluster.
        \textbf{Claims:}
        \ctext[RGB]{253,253,150}{1. An apparatus, comprising: at least one processor; and at least one memory including computer program code, wherein the at least one memory and the computer program code are configured to, with the at least one processor, cause the apparatus to at least...} &  An apparatus for achieving phase resolving in single-frequency networks. \ctext[RGB]{253,253,150}{The device includes a memory and a program code that causes the device to receive signals from multiple different types of sources, each of which has a quality metrics associated with it}. A cluster is formed wherein information regarding the selected point is given and an indication about the other points is given. In further embodiments, the device selects a particular point to achieve better signal quality than standard \ctext[RGB]{255, 179, 71}{bribs}. & LongT5 \\ 
\hline
\end{tabular}
}
\begin{tablenotes}
    \scriptsize 
    \item Note:
 Phrases in \ctext[RGB]{253,253,150}{yellow} are copied from Abstract, while phrases in \ctext[RGB]{255, 179, 71}{orange} are excerpted from Claims.
\end{tablenotes}
\end{table*}

\end{document}